# Reservoir Computing based Neural Image Filters

Samiran Ganguly
*Charles L. Brown Dept. of Electrical and Computer Engineering*
*University of Virginia*
Charlottesville, VA 22904
sganguly@virginia.edu

Yunfei Gu
*Charles L. Brown Dept. of Electrical and Computer Engineering*
*University of Virginia*
Charlottesville, VA 22904

Yunkun Xie
*Charles L. Brown Dept. of Electrical and Computer Engineering*
*University of Virginia*
Charlottesville, VA 22904

Mircea R. Stan
*Charles L. Brown Dept. of Electrical and Computer Engineering*
*University of Virginia*
Charlottesville, VA 22904

Avik W. Ghosh
*Charles L. Brown Dept. of Electrical and Computer Engineering*
*University of Virginia*
Charlottesville, VA 22904

Nibir K. Dhar
*US Army Night Vision & Electronic Sensors Directorate*
Fort Belvoir, VA 22060

*Abstract*— Clean images are an important requirement for machine vision systems to recognize visual features correctly. However, the environment, optics, electronics of the physical imaging systems can introduce extreme distortions and noise in the acquired images. In this work, we explore the use of reservoir computing, a dynamical neural network model inspired from biological systems, in creating dynamic image filtering systems that extracts signal from noise using inverse modeling. We discuss the possibility of implementing these networks in hardware close to the sensors.

## I. INTRODUCTION

Electronics industry in 21st century is being driven by the techno-economic trend of pervasive computing embedded in smart devices, and availability of high-speed data networks. There is a huge demand for self-driving automotives, airborne platforms or drones in both security as well as private sector, particularly logistics, geo-exploration, weather monitoring, and disaster recovery, smart homes etc. which are now collectively being called the "Internet of Things" (IoTs).

Many such IoT devices involve machine vision, where a computer is expected to automatically acquire images from a camera, process it, and then take action based on the image content. To perform this task successfully, it is critical that the acquired images are "clean" and ideally only include targeted features in the image frame.

However, many such IoT devices are expected to work in varying environments, lighting conditions, visibility etc. The imaging sensors themselves can be important sources of noise and distortions, arising from the camera optics, as well electrical response of the detector material and circuitry, e.g. IR sensors and bolometers are highly sensitive to temperatures. It is imperative that high performance, compact non-linear imaging filters be developed for these applications that can adapt to the challenging task of uncontrolled environment.

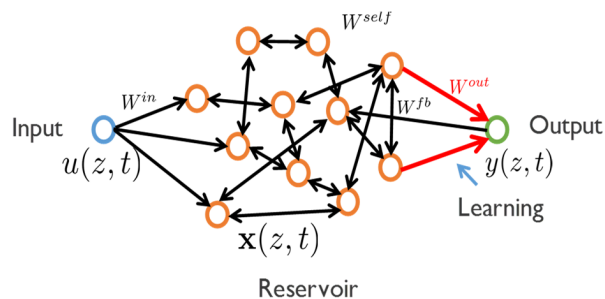

Fig. 1. General schematic of a reservoir computer. Reservoir is composed of a collection of weakly coupled analog "leaky-integrate-and-fire" (LIF) neurons, connected recurrently, i.e. with feedbacks. One of the nodes acts as the input with state given by $u(z,t)$, and one of them as the output $y(z,t)$, while the rest of the network's state is represented by $x(z,t)$. The synaptic weights are represented by various matrices $W$ as shown. The only synaptic weight adjusted is $W^{out}$, typically adjusted using a linear regression technique such as Weiner-Hopf or Tikhonov regularization.

Deep Neural Network architectures, in their myriad forms, including deep Convolutional Neural Networks (CNNs) [1], Restricted Boltzmann Machines (RBMs) [2], and Long-Short Term Memory (LSTM) [3] have made huge advances in practical implementation with multiple machine learning libraries available from all major software vendors as well as academia. The primary reason such advances have been possible is due to aggressive scaling of transistor (Moore's Law) and development of sophisticated cloud computing infrastructure that has brought powerful GPU based clusters outside of supercomputing center business into everyday consumer market. High Performance Computing (HPC) is now accessible at a drastically reduced price point.

However, these solutions necessarily involve huge energy consumption and there are fundamental application scaling challenges due to the near-end of Moore's Law style transistor scaling [4]. In absence of extreme computing capabilities at each individual IoT nodes, such devices will depend on a background high speed data network to leverage neural networks

This work was supported by the NSF I/UCRC on Multi-functional Integrated System Technology (MIST) Center IIP-1439644, IIP-1738752 and IIP-1439680 implemented in the cloud, similar to services like Siri, Alexa, and Google Assistant. However, this opens up a challenging task of cyber security and possibilities of physical network disruption [5], which can be debilitating in a mission critical and remote location applications.

Therefore, there is a need to explore neural network architectures, which are better suited for applications with strict size, weight, and power (SWaP) limitations.

In this work, we have presented neuro-adaptive filters capable of dealing with "spatio-temporal" signal, e.g. video data, developed using a class of recurrent networks called Reservoir Computers that employ *simple learning techniques without any backpropagation, and therefore are straightforward to implement on an embedded processor on-board with IoT devices as a system-on-chip (SoC),* reducing the computation and training load significantly. We hope that this work will spark an interest in exploring the applications of less explored neural architectures such as reservoir computers in IoT space and neuro-adaptive signal processing.

## II. RESERVOIR COMPUTING FUNDAMENTALS

### A. Computing Using Dynamical Systems

Reservoir Computers (RC), in two versions – the Echo-State Networks (ESNs) [6] and Liquid State Machines (LSMs) [7] are examples of dynamical systems used for computation. These networks are particularly suited for multi-dimensional classifications of signals, making them particularly suited for time varying signal (time being one of the dimensions) classification [8]. Being able to tune networks to various dynamical time scale, allows for processing of signals of varying bandwidth [9]. The rest of this section describes the basics of RC, focusing on the ESNs.

### B. Reservoir Design

Central principle of reservoir computing is the high degree of recurrence in the collection of the neurons (fig.1). Recurrence in this context means structural feedbacks giving rise to memory states in the dynamics of the reservoir [10]. These feedbacks allow an input signal $u(z,t)$ at $t = \tau$ to persist in the future, i.e. for $t > \tau$, due to finite speed of signal propagation within the nodes of the reservoir. This turns the reservoir into a *temporal correlator* wherein the patterns of the signal in time dimension can be stored and classified.

As mentioned before, RC has been developed in two flavors: a) ESN – the neurons composing the reservoir are stochastic leaky-integrate-and-fire (LIF) type neurons, which can accumulate the signals at its input and if this accumulation build up to a certain threshold within a certain timescale, the neuron fires [11]. The output $y(z,t)$ is a linear readout of the collective reservoir state/activations given by $x(z,t)$. b) LSM – the neurons have spiky activations, e.g. Fitzhugh-Nagumo [12] or Izhikevich [13] neuron models. The information is encoded in temporal distance between the consecutive spikes or pulses. A deep neural network rather than a linear sampler does the readout, in case of LSMs.

Both the models use similar principles of operation, in the case of ESN, the network states are analog signals, while in case of LSM are spike trains. The learning in the case of ESN is a straightforward one-pass linear regression, and as a result, we have chosen it for our work, which deals with noisy analog signals rather than pulse trains.

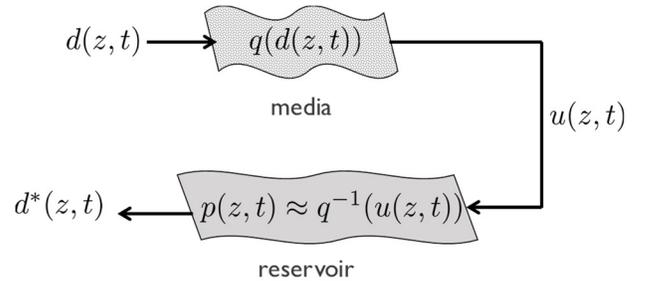

Fig. 2. Filtering by Inverse Modeling. The detecting media (environment + detector material and electronics) introduces distortion to the signal $d(z,t)$ through an unknown non-linear function $q(d(z,t))$ to produce $u(z,t)$. The reservoir is trained to reconstruct the original signal by producing the inverse function $q^{-1}(u(z,t))$ during supervised training. After training, the network works as a dynamic neural-filter that undoes the effect of the media to recover the original signal.

### C. Reservoir Dynamics and Training

The reservoir's dynamics is given by the equation:
$$\frac{dx}{dt} = -\gamma x + \kappa \tanh(W^{self}x + W^{in}u + W^{fb}y) \quad (1)$$
$$y = W^{out}x \quad (2)$$

Where, $x$ is the collective dynamical state of the network, $u, y$ are the input and output of the network, the various matrices $W^{self}, W^{in}, W^{fb}, W^{out}$ are the synaptic weights between the various components of the system, shown in fig. 1. It can be shown using standard theory of differential equations that the above 1st order differential equation (eq. 1) is a non-linear initial value problem (IVP) with memory states, due to its self-interaction arising from structural recurrence given by $W^{self}$ [14].

It is critical to note that the self-interaction is kept weak otherwise the dynamics in the reservoir can turn chaotic (analogous to positive feedback). This is detrimental to the "echo-states" in the network, where the signature of past activations are supposed to persist in a fading sense (analogous to negative feedback). Without the echo-states the network does not work as intended. Empirically, it has been found that scaling the spectral radius of $\rho(W^{self}) < 1$ helps in ensuring echo-states [15]. Reservoir computers can also be viewed as special case of Markov chains, which explains their ability to model temporal correlations [16].

The readout (eq. 2) is a linear combination of the reservoir states at any given time, given by $W^{out}$, and can be trained using a linear regression technique.

## III. DYNAMIC FILTERING

In this section, we discuss two filtering tasks using ESNs. We first discuss the general idea behind using ESNs for filtering and then discuss two particular filtering tasks.

### A. Filtering by Inverse Modeling

The central idea behind dynamic filtering using ESN is inverse modeling of a time series generator [17]. A media,

which in our case is the environment + detector optics and electronics, which lies between the object being sensed and the overall sensor system output, can introduce all manners of amplitude and phase noise, and non-linear response causing image aberrations, which we can collectively call distortions.

These distortions can be thought of as the response of the media given by an *unknown* functional $q()$, mapping an input signal function $d(z,t)$ to a distorted output function $u(z,t) = q(d(z,t))$. However, we assume that we can access the media and generate the teacher data pairs: $\{d(z,t)\ u(z,t)\}$. We can then train an ESN to reverse generate this pair, i.e. generate $d^*(z,t)$ from $u(z,t)$ and minimize the error: $||d(z,t) - d^*(z,t)||$ by adjusting the readout given by the matrix $W^{out}$ (fig.1) using algorithms such as Weiner-Hopf [18] or Tikhonov regularization [19] (also called ridge regression). This effectively converts the RC into an *inverse model* of the media response functional $q$: $p() = q^{-1}()$ (fig.2).

In this work, we created our own set of equations to create the media response functional $q$ and then use it to train and test as discussed next.

### B. 1-D Non-Linear Distortion Filtering

For a 1-D case, we use a non-linear distortion functional given by:

$$q(t) = \sum_k a_k \left[\sum_l b_l d(t-l)\right]^k + c_k r_n \quad (3)$$

Where $k$ and $l$ are integers, $a, b, c$ are chosen system parameters and $r_n$ is a Gaussian distributed random number generator.

In this case, the spatial dimension of the signal is 0, i.e. it is a scalar signal and can stand in for signals such as audio, bio-physical signals etc.

### C. 2-D Video Filtering

In this case, the signal $d(z,t)$ has spatial dimension of 2, i.e. $z = (\hat{x}, \hat{y})$. Therefore, the data stream is a video.

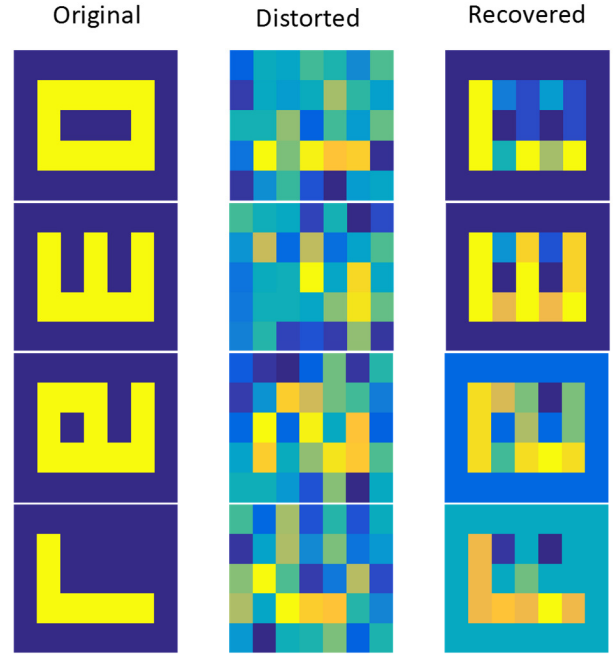

Fig. 4. Filtering of 2-D frames. It can be seen that even after severe distortion of the original signal, the reservoir can recover the image and the original glyphs can be identified. Creating a deep readout, either through a hierarchical reservoir or using a deep readout network, such as a CNN may enable higher fidelity of recovery.

We create a set of image glyphs that we then stack together to form a video stream. Each of the pixel in the video stream is distorted by adding noise, generated by a Gaussian random number generator. The distortion functional is given by:

$$q(z,t) = a_t d(z,t) + c_t r_n \quad (4)$$

It should be noted that in this case, we have not included any non-linearity in the system, to keep the ESN sizes small, as distortions that are more complex require larger networks to filter.

In next section, we discuss the results of neural filtering on both 1-D non-linearly distorted data, as well as 2-D noisy video.

## IV. RESULTS AND DISCUSSIONS

Reservoir computing, being an example of computing using a dynamical system throws up a huge amount of richness in their behavior, and are sensitive to the particular choice of system parameters used. To present a coherent picture that demonstrates the filtering tasks, we have chosen to use fixed and consistent sets of parameters for the two different tasks. We have not attempted to optimize these parameters to get the most efficient performance, merely to find a reasonably useable set.

In the case of 1-D filtering, we use a relatively small network of size N = 20, i.e. composed of 20 nodes. We generate a bit stream of data and pass it through the 1-D distortion functional (eq. 3) to generate the pair $\{d(t)\ u(t)\}$. We split a part of this tuple to generate training data (2000 samples) and the test data (2000 samples).

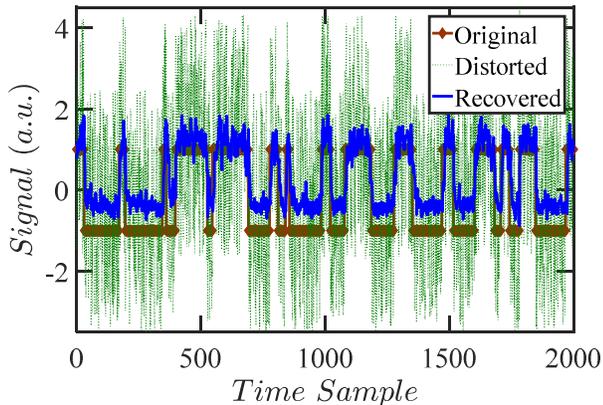

Fig.3. Filtering of 1-D signal. A binary (between -1 and 1) data stream is distorted and recovered to a high degree of fidelity by using a small reservoir (N = 20). The reservoir is taught the correlation between the original signal and distorted data, and the readout is trained to reconstruct/recover the original signal.

It can be seen from fig. 3 that the network can reconstruct or recover the original signal even from severely distorted signal, purely using network activations and a linear readout. The network captures the line shapes and the transitions successfully, which means that all the symbols embedded in the distorted signal are identified. An extra layer of smoothing and shifting circuitry can be combined with this readout to recover the signal fully.

In our experience of running the network over many examples, we have observed between $90-100\%$ signal recovery depending on the complexity of the distortion functional, reservoir/network size, and training sample size with a mean recovery rate $\approx 94\%$ for non-hierarchical reservoirs (i.e. only a simple linear readout from a single reservoir). The signal recovery improves with larger reservoirs, though it seems to saturate as a function of size, depending on the difficulty of the problem.

We then turn our attention to the task of video filtering. In this case, we have used a 500-node network and the input and output nodes are 3-D vectors of the image frame size ($L \times W$) + number of time samples ($T$), i.e. $d \equiv \{T, L, W\}$. Our strategy of generating the teaching and test data remains the same. We first generate a video stream, which is an array of 2-D frames composed of glyphs shown in fig. 4. We use the data stream consisting of multiple glyphs to train the network (3000 frames) and generate the distorted output, which then forms the input to the network.

Fig. 4 shows the example recovery of the images in four different frames of the video data. It can be seen that again the network is successful in recovering the glyphs to a reasonable extent from severely distorted data. Similar to the 1-D data, it is possible to combine the reservoir with a deep neural network based readout or multiple stacked reservoirs (deep ESNs) to fully recover the signal. Nevertheless, the network in itself does the task of image recovery to an impressive extent. We have not yet studied the statistics of video signal recovery task by ESNs to the extent we have for the 1-D signal filtering task, we expect the accuracy rates to lie in the same range.

ESN based networks can also open up the space for optimization of an image recognition system composed of deep neural networks such as CNN by reducing the complexity of feature maps through noise and distortion reduction. Novel architectures combining reservoirs with convolutional filters as read-outs could enable efficient low depth networks that outperform conventional CNNs of similar network size, due to embedded temporal inferencing built-in with such architectures.

It should be noted that we have used small reservoirs to demonstrate the capabilities of ESNs to perform these tasks with high computational efficiency. To embed the network in a hardware based signal processor, it is necessary to use small optimized networks to deal with real-time data of high frame-rates.

It should be further noted that even though the training used here is supervised, there has been advances in unsupervised methods [20], which will allow development of on-line real time smart image filters.

## V. HARDWARE NEURAL ROICs

We conclude this paper with a discussion on possibility of embedding ESN based dynamic filters in the read-out ICs (ROICs) or Signal Processors on board with the sensors.

Field Programmable Gate Arrays (FPGAs) have been a platform of choice in digital communication and signal processing for implementation of filtering, coding, and transceiver functionalities due to their programmability [21]. Neural networks are inherently dynamic in nature, and are increasingly being implemented on them [22]. FPGAs can be directly combined with a camera ROIC for implementation of neuro-adaptive signal processing capabilities within the same device with built-in data network independence and resilience.

FPGAs consist of a large number of look-up-table (LUT) based logic blocks [23] that can implement any basic Boolean operation, with more complex digital designs built out of clusters of smaller LUT based gates. As such, FPGA synthesis can be used to implement fast parallel arrays of mathematical operations that are necessary to implement neural networks, such as dot-products and sums built from smaller Boolean gates. These FPGAs can be used as "linear algebra" accelerators [24] and in that respect work as a more efficient accelerator than GPU based implementation.

Increasingly, there is a trend to implement a learned neural network model on an FPGA as a dataflow architecture directly from a high level code [25] (say python tensorflow). This has shown a lot of promise and can open up a pathway for fast implementation and dynamic programming of FPGAs with continually updated neural network models. These advances have provided new design methodologies for embedding energy-efficiency aware neural network models directly into a camera ROIC.

However, FPGAs still suffer from the issues of limits to transistor scaling, and an inherent mismatch between the mathematical operations of neural networks (linear algebra of continuous vector spaces) and the LUT primitives (linear algebra of discrete vector spaces). Therefore, it is reasonable to expect a limit to performance and efficiency from these implementations.

There is an increasing interest in devices and computational capabilities offered by emerging nano-materials technology such as memristors and spintronics, which embed within themselves the primitives for efficient neural network operations due their inherent physics, such as thresholding, built-in memory, and true stochasticity.

Current capabilities in advanced fabrication of nano-materials have now opened up the possibility of embedding complex logic and processing capabilities using these nano-materials with a combination of conventional CMOS platform to build a System-on-Chip (SoC). Elsewhere [26, 27, 28], we have demonstrated that it is possible to fabricate the hardware primitives that will be necessary to build such neural networks in hardware today.

These hardware primitives implement the basic functionality of LIF neurons in a compact footprint, i.e. using just 3-4 components, including these nano-materials based devices with a combination of conventional CMOS based invertors and buffers. Small footprint provides high energy-efficiency, density of fabrication, and ultra-scalability not offered by conventional CMOS only platform. With better control over fabrication and variability of these emerging nano-materials, we can eventually expect large-scale adoption of hardware-based neural signal processors embedded in-situ with sensors.

ACKNOWLEDGMENT

This work was supported by the NSF I/UCRC on Multi-functional Integrated System Technology (MIST) Center IIP-1439644, IIP-1738752 and IIP-1439680.